%% file: main.tex
\documentclass[letterpaper]{article} % DO NOT CHANGE THIS
\usepackage{aaai25}  % DO NOT CHANGE THIS
\usepackage{times}  % DO NOT CHANGE THIS
\usepackage{helvet}  % DO NOT CHANGE THIS
\usepackage{courier}  % DO NOT CHANGE THIS
\usepackage[hyphens]{url}  % DO NOT CHANGE THIS
\usepackage{graphicx} % DO NOT CHANGE THIS
\usepackage{natbib}  % DO NOT CHANGE THIS AND DO NOT ADD ANY OPTIONS TO IT
\usepackage{caption} % DO NOT CHANGE THIS AND DO NOT ADD ANY OPTIONS TO IT
\usepackage{algorithm}
\usepackage{algorithmic}
\usepackage{amsmath}
\usepackage{booktabs}
\usepackage{xcolor}

\newcommand{\red}[1]{\textcolor[rgb]{1,0,0}{#1}}

\newcommand{\ignorethis}[1]{}
\usepackage{newfloat}
\usepackage{listings}
\DeclareCaptionStyle{ruled}{labelfont=normalfont,labelsep=colon,strut=off} % DO NOT CHANGE THIS
\floatstyle{ruled}
\newfloat{listing}{tb}{lst}{}
\floatname{listing}{Listing}
\title{GazeTrack: High-Precision Eye Tracking Based on Regularization and Spatial Computing}

\author{
    \textbf{\fontsize{12pt}{14pt}\selectfont Xiaoyin Yang}\\
    \textnormal{\fontsize{10pt}{12pt}\selectfont Dalian University of Technology}\\
    \textnormal{\fontsize{10pt}{12pt}\selectfont tooyoungalex@outlook.com}
}

\usepackage{bibentry}
\begin{document}

\maketitle

\begin{abstract}
  Eye tracking has become increasingly important in virtual and augmented reality applications; however, the current gaze accuracy falls short of meeting the requirements for spatial computing. We designed a gaze collection framework and utilized high-precision equipment to gather the first precise benchmark dataset, GazeTrack, encompassing diverse ethnicities, ages, and visual acuity conditions for pupil localization and gaze tracking. We propose a novel shape error regularization method to constrain pupil ellipse fitting and train on open-source datasets, enhancing semantic segmentation and pupil position prediction accuracy. Additionally, we invent a novel coordinate transformation method similar to paper unfolding to accurately predict gaze vectors on the GazeTrack dataset. Finally, we built a gaze vector generation model that achieves reduced gaze angle error with lower computational complexity compared to other methods.Please refer to our project page for more details: https://github.com/----(please refer to other supplementary materials).
\end{abstract}


\input{sections/introduction}
\input{sections/relatedwork}

\input{sections/GazeTrack}
\input{sections/dataset}

\input{sections/experiments}

\input{sections/conclusion}

\clearpage
\bibliography{aaai25}
\clearpage

\input{sections/checklist}
\end{document}

%% file: sections/introduction.tex
\begin{figure*}[th]
	\centering
	\includegraphics [width=1.0\linewidth]{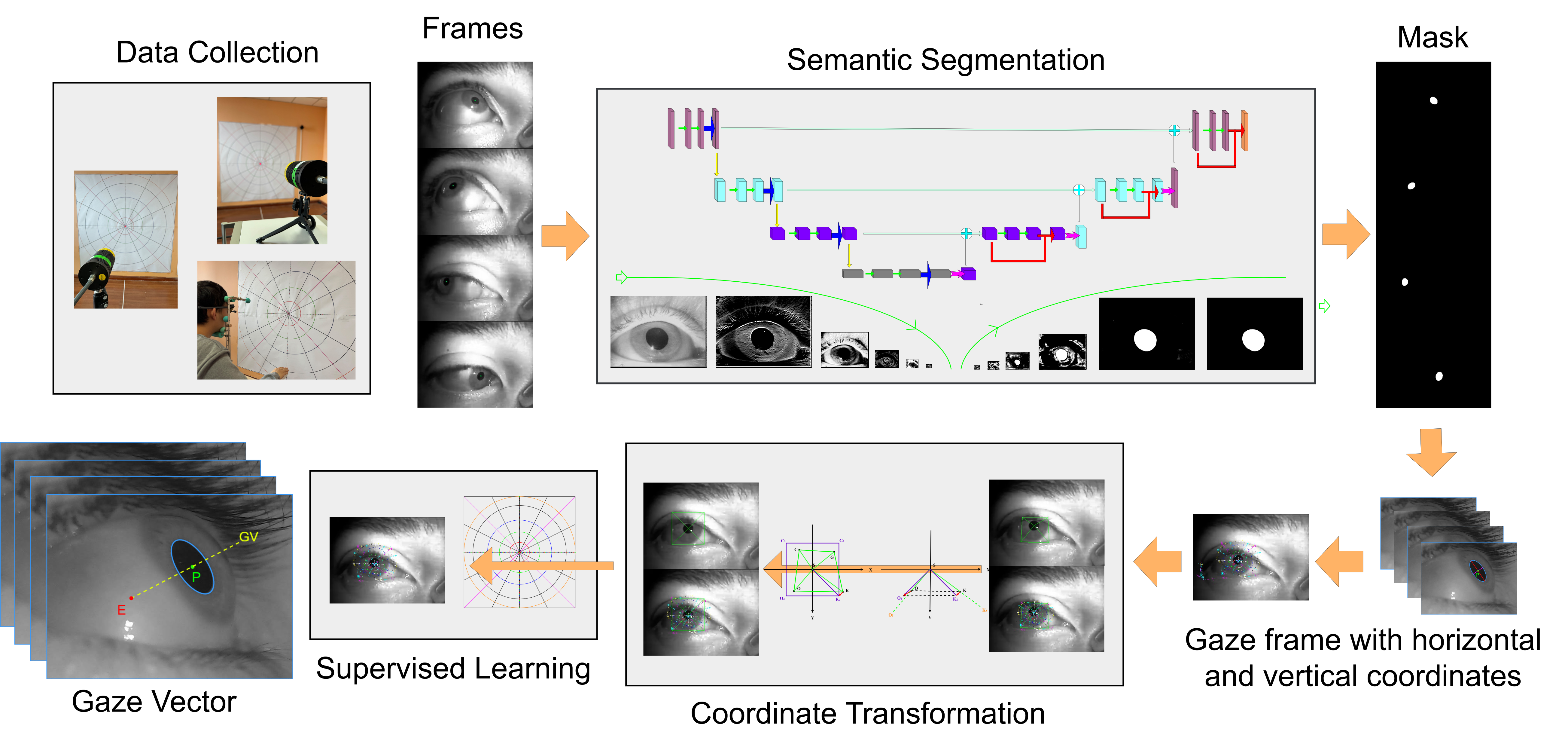}
	\caption{The steps of the GazeTrack}
	\label{fig:GazeTrackPipeline}
\end{figure*}
\section{Introduction}

The rise of the era of spatial computing\cite{balakrishnan2021interaction} has ushered in unprecedented development opportunities for human-computer interaction (HCI) methods. As an important means of HCI, eye-tracking technology has gradually become a hot topic in both research and applications. Eye-tracking technology enables more intelligent and intuitive interaction by tracking the user's eye movements, significantly enhancing the user experience. Gaze vector generation technology based on spatial computing represents a new direction in the development of eye-tracking technology.

On one hand, this technology allows for precise monitoring and analysis of user gaze behavior, providing personalized interaction experiences for users. On the other hand, it can be applied in fields such as virtual reality\cite{anthes2016state} and intelligent driving\cite{wu2020motionnet}, offering new ideas and methods for the development of related technologies. This is of great significance for advancing HCI technology.

However, there are numerous challenges in achieving gaze vector generation. Firstly, eye-tracking involves processing and analyzing eye images, which are often affected by environmental lighting and rapid eye movements, resulting in poor image quality. Secondly, the complex shape of the eye makes it difficult for traditional image processing methods to accurately extract eye contours. Accurate data is required for gaze vector generation, but existing data precision is insufficient. Additionally, current hardware struggles to meet the computational complexity requirements of gaze vector generation, which demands low latency and high frame rates.

To address these challenges, our research employs techniques such as semantic segmentation, ellipse fitting, and coordinate transformation to accurately process and analyze eye images, ultimately generating gaze vectors.

In particular, we make the following contributions in this work:
\begin{itemize}
    % \vspace{-0.1cm}
    \item We collected and processed a highly precise gaze dataset and method, GazeTrack, comprising data from various ages, genders, and ethnicities. We validated the optimal performance of our algorithm on this dataset and third-party datasets.
    % \vspace{-0.1cm}
    \item We innovated the semantic segmentation model U-ResAtt, which effectively separates the eye regions from the surrounding environment in eye images, thereby improving the accuracy of eye contour extraction.
    % \vspace{-0.1cm}
    \item We proposed a shape prior regularization constraint for pupil ellipse fitting, achieving more precise pupil coordinates.
    % \vspace{-0.1cm}
    \item We created a gaze coordinate transformation method, CoordTransNet, suitable for gaze data collected from various angles.
    % \vspace{-0.1cm}
    \item We developed the gaze vector generation model, GVnet, using a sliding window technique. This model can generate accurate gaze vectors from a single image after training.
\end{itemize}

\textbf{limitations} Eye-tracking devices typically need to be worn on the head, and vigorous head movements may introduce interference to eye-tracking data, leading to distortion and reduced accuracy.

%% file: sections/relatedwork.tex
\section{Related Work}

% \section{Background and Related Work}
\label{sec:RW}
% \vspace{-0.2cm}
% \subsection{Background}\label{sec:back}
% \vspace{-0.2cm}

Semantic segmentation, image deformation, coordinate transformation, gaze estimation, and spatial computation are important research areas involving computer vision, human-computer interaction, and artificial intelligence, among other fields. Multiple teams are dedicated to research and development in these domains.

The U-Net proposed by O. Ronneberger\cite{ronneberger2015u} combines fully convolutional networks (FCNs) with autoencoder structures to achieve end-to-end semantic segmentation. With a U-shaped architecture consisting of symmetric encoder and decoder parts and skip connections to fuse low-level and high-level features, it effectively enhances segmentation accuracy and reduces the risk of overfitting during model training. U-Net has shown outstanding performance in neural structure segmentation tasks in electron microscopy stack images and emerged victorious in the ISBI cell tracking challenge of that year.

W. Fuhl\cite{fuhl2016eyes}, focusing on semantic segmentation of the eye region, proposed a method for universal eye tracking that includes eyelid recognition and pupil estimation. This approach achieved up to a 40\% performance improvement in outdoor driving scenarios, demonstrating real-time operation and high accuracy. The team, while maintaining model performance and real-time capability, introduced a new method\cite{fuhl2020training} that replaces convolutional layers with decision tree layers, significantly reducing computational complexity. They also compared the effectiveness of using convolutional, binary, and decision tree layers in neural networks for estimating pupil contours\cite{fuhl2020training}, proving that real-time performance can be maintained while improving the accuracy and robustness of pupil contour estimation. The team combined Haar-like features\cite{fuhl20211000} and statistical learning techniques to propose a pupil segmentation method suitable for mobile eye tracking and high-speed gaze tracking.

Mirikharaji Z and Hamarneh G optimized the performance of fully convolutional networks (FCNs) in skin lesion segmentation tasks by introducing Star Shape priors\cite{mirikharaji2018star}, achieving leading experimental results. Inspired by this, Han et al. proposed a robust pupil center detection method based on CNNs by introducing shape priors\cite{han2020noise}, enhancing robustness against noise. Schnieders\cite{schnieders2010reconstruction} reconstructed eye models from single images to calculate binocular vectors.

Microsoft's Erroll Wood team\cite{wood2016learning} utilizes computer vision to infer gaze points or attentional focuses of human eyes, improving user interface design, monitoring systems, etc. Google has patented eye tracking for autonomous driving based on understanding drivers' attention and intentions from gaze direction, enhancing its autonomous driving system with eye tracking technology to make it smarter and safer. NVIDIA introduced the NVGaze dataset\cite{kim2019nvgaze} for low-latency near-eye gaze estimation based on anatomical information, including variations in facial shape, gaze direction, pupil and iris, skin color, and external conditions. They also collected a real-world dataset from 35 subjects. Utilizing these datasets, they trained neural networks achieving sub-millisecond latency, with gaze estimation error of 2.06 degrees (±0.44 degrees).

Dierkes\cite{dierkes2018novel}, Swirski\cite{swirski2013fully}, and others used multiple-view images to obtain geometric information of the eye, estimating eye shape and position for gaze prediction through mathematical modeling and optimization algorithms. Otmar Hilliges\cite{zhang2020eth} created and open-sourced the ETH-XGaze dataset and proposed a new loss function for unsupervised learning. Jean Marc Odobez\cite{siegfried2022robust} employed task attention priors and multimodal information for gaze point estimation. Sugano\cite{lu2015gaze}\cite{zhang2017mpiigaze} collected the MPIIGaze dataset, comprising 213,659 full-face images of users during daily laptop use, along with corresponding real gaze positions. Data were collected using an experience sampling method to ensure continuity of gaze and head pose, considering real changes in eye appearance and lighting.

Compared to previous methods, our approach improves the accuracy of gaze vector generation, reduces computational complexity, and is applicable to eye data collected from various angles, making high-precision spatial computation more practical.

%% file: sections/GazeTrack.tex
\section{Eye Tracking Methods Based on Spatial Computing (GazeTrack)}

Firstly, we developed a U-ResAtt model based on U-Net, incorporating attention mechanisms and residual connections. We combined the binary cross-entropy with the Ellipse Fit Error (L\(_{\text{EFE}}\)) regularization method to effectively constrain the loss, thereby reducing errors caused by reflections, occlusions, and rapid eye movements, and improving the accuracy and robustness of pupil localization. Secondly, using the concept of image deformation and linear interpolation, we created a coordinate transformation method that adjusts the gaze distribution maps collected from subjects into a standard shape. Finally, we built a gaze generation model, training it using a sliding window as input. Below, I will describe the components of this method in detail.

\subsection{Accurate Pupil Segmentation with Error Regularization(U-ResAtt)} 
{\textbf{Model Overview} To achieve precise semantic segmentation of the pupil, we incorporated attention mechanisms and residual connections\cite{he2016deep} into the U-Net model. This addition aims to enhance segmentation accuracy in challenging conditions such as low light, reflections, and rapid eye movements. By reducing the original number of layers, we addressed real-time processing constraints. The enhanced U-ResAtt model structure is depicted in Figure \ref{fig:u-resatt_model}.
\begin{figure}[h]
	\centering
	\includegraphics [width=1.0\linewidth]{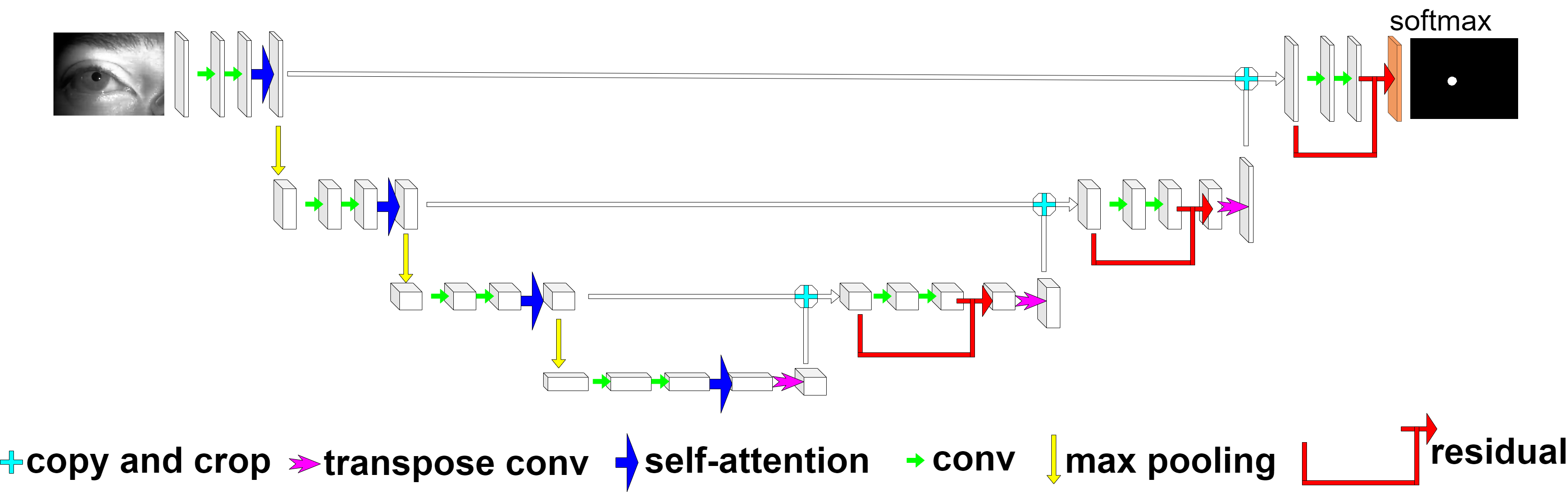}
	
	\caption{
      U-ResAtt model
	}
	\label{fig:u-resatt_model}
\end{figure}

{\textbf{Binary Cross Entropy} Among the numerous possibilities, the Binary Cross-Entropy Loss ($L_{BCE}$) function is one of the commonly used loss functions in image segmentation\cite{yeung2022unified} tasks. For binary classification problems, this loss function measures the difference between the probability distribution predicted by the network and the true label's probability distribution. The function is computed as follows:

\begin{equation}
    L_{BCE} = -\sum_{i \in I}^N \sum_{p \in \Omega} y_{ip} \log S_{ip} + (1 - y_{ip}) \log (1 - S_{ip})
    \label{eq:bce_loss}
\end{equation}

where $i$ denotes the $i$-th batch among $N$ images, $\Omega$ represents the pixel space of the image, $y$ represents the Ground Truth (GT) image, and $S$ represents the Sigmoid output of the CNN. In the case of pupil segmentation, there are only two categories in the GT: 0 representing background pixels and 1 representing pixels composing the pupil. Therefore, $y_{ip}$ is either 0 or 1. $\log$ denotes the natural logarithm. The negative log-likelihood form of the binary cross-entropy loss function ensures the maximization of the probability of the correct class and minimization of the probability of the incorrect class.

In image segmentation tasks, the binary cross-entropy loss function for the entire image is defined as the sum or average of the loss functions for all pixels. By minimizing this loss function, the network can learn to correctly segment the image into two parts: the target and background.

{\textbf{Ellipse Fit Error} Despite the satisfactory performance of U-Net using Binary Cross-Entropy (BCE) as the loss function in many image segmentation applications, we demonstrates that BCE alone is insufficient. To avoid issues such as motion blur and reflection, which affect pupil extraction, it is necessary to add a regularization term to the loss function. This term, named Ellipse Fit Error ($L_{EFE}$), is calculated based on the shape of the detected object and multiplied by a weight parameter added to the loss. 

The Ellipse Fit Error loss is added as a regularization term to the pixel-level loss function to improve pupil semantic segmentation. Since the pupil in a 2D image is an ellipse, the goal is for the predicted pupil segmentation shape to be a convex polygon tending toward an ellipse rather than just a pixel edge segmentation, perfectly matching the pupil position. Any deviation between the predicted pupil and the ground truth (GT) ellipse will be added as an extra regularization term to the pixel-level binary cross-entropy loss function, enforcing the model to produce elliptical semantic inferences.

\begin{figure}[h]
	\centering
	\includegraphics [width=0.5\linewidth]{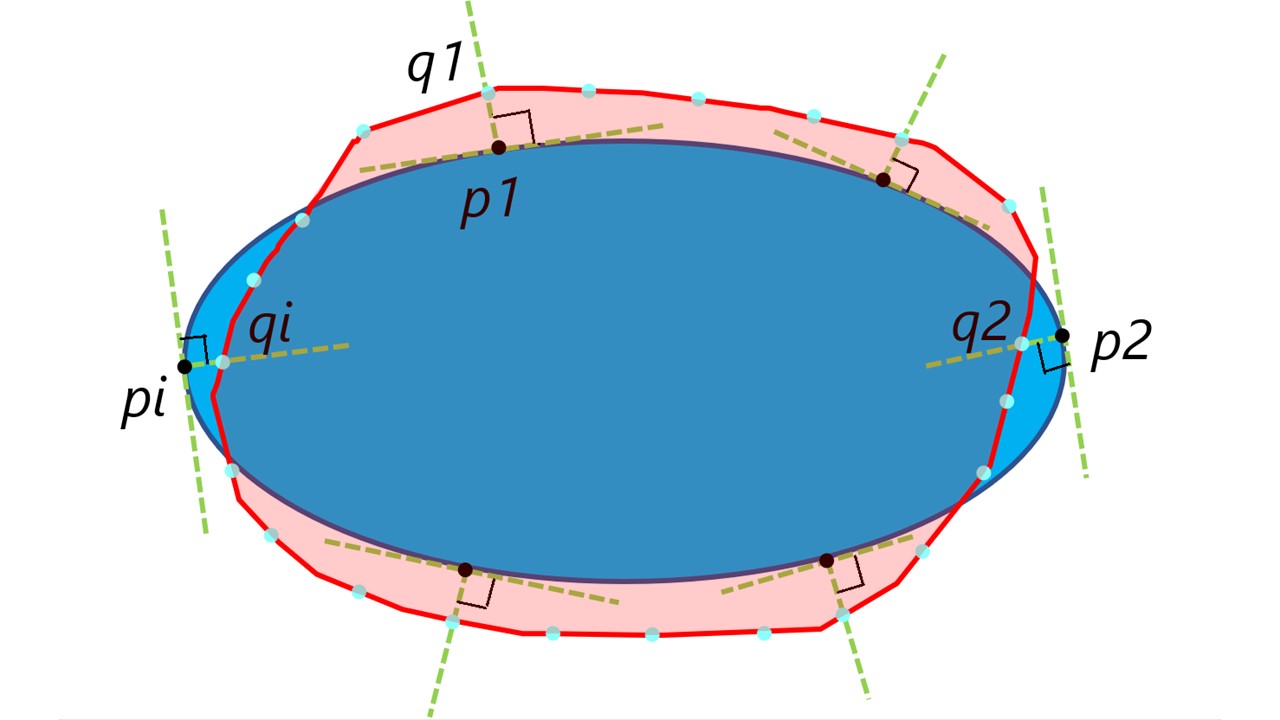}
	
	\caption{
      Ellipse fit error
	}
	\label{fig:ellipse_fit_error}
\end{figure}

Figure \ref{fig:ellipse_fit_error} illustrates how the proposed regularization term $L_{EFE}$ is calculated. First, the ellipse representing the GT pupil boundary is computed. Next, the Edge Drawing algorithm\cite{topal2012edge} is used to calculate the boundary pixels of the predicted pupil spot (cyan points $q$ in Figure \ref{fig:ellipse_fit_error}). For each pixel on the predicted pupil boundary, the nearest point on the GT ellipse (blue points on the ellipse boundary in Figure \ref{fig:ellipse_fit_error}) is computed. The Euclidean distance between $p$ and $q$ is then added to $L_{EFE}$. Finally, $L_{EFE}$ is added as a regularization term to the loss function as described in Equation \ref{eq:total_loss}:

\begin{equation}
    L = \alpha L_{BCE} + \beta L_{EFE}
    \label{eq:total_loss}
\end{equation}

The idea behind the Ellipse Fit Error ($L_{EFE}$) regularization term is as follows: If a point on the detected pupil spot boundary is outside the GT (such as $q1$), this additional loss term will push that point towards $p1$ on the GT boundary. Conversely, if a point on the detected pupil spot boundary is inside the GT (such as $q2$), this additional loss term will push $q2$ towards $p2$ on the GT boundary. When all boundary pixels of the detected pupil spot lie on the GT boundary, $L_{EFE}$ tends to 0, which is the desired outcome of this study.

In mathematical terms, the ellipse equation representing the GT pupil boundary is given by:

\begin{equation}
    Ax^2 + Bxy + Cy^2 + Dx + Ey + F = 0
    \label{eq:ellipse_eq}
\end{equation}

This ellipse equation can be converted to a parametric form as shown in Equation \ref{eq:parametric_ellipse}:

\begin{equation}
    \left(\frac{x}{\alpha}\right)^2 + \left(\frac{y}{\beta}\right)^2 = r^2
    \label{eq:parametric_ellipse}
\end{equation}

Given the predicted pupil spot on the boundary $q = (x, y)$, the goal is to compute the nearest point $p$ on the ellipse boundary. If $p$ is represented in polar coordinates as in Equation \ref{eq:polar_coords}, the tangent of $\vec{p}$ must be perpendicular to the line $\vec{q} - \vec{p}(\theta)$:

\begin{equation}
    \vec{p} = \left( r\alpha \cos(\theta), r\beta \sin(\theta) \right)^T
    \label{eq:polar_coords}
\end{equation}

Equation \ref{eq:polar_coords} can now be iteratively solved for $\theta$ using Equation \ref{eq:iterative_solution}:

\begin{equation}
    (\vec{q} - \vec{p}(\theta)) \cdot \vec{p}'(\theta) = 0
    \label{eq:iterative_solution}
\end{equation}

It is recommended to use $\tan^{-1}\left(\frac{\alpha y}{\beta x}\right)$ as the initial value for $\theta$. This system of equations can be solved numerically using the Newton-Raphson method, and according to Cakir \& Topal\cite{cakir2018euclidean}, convergence for each point requires two to three iterations. Once the nearest GT ellipse boundary pixel $p$ is found for each detected pupil pixel $q$, the distance between the two contours in Euclidean space is calculated using Equation \ref{eq:efe_distance}:

\begin{equation}
    L_{EFE} = \sum_{\substack{p \in e_{GT}\\ q \in e_P}} \|p - q\|
    \label{eq:efe_distance}
\end{equation}

where $e_{GT}$ represents the edge pixels of the GT pupil and $e_P$ represents the edge pixels of the detected pupil. As $L_{EFE}$ approaches 0, the detected pupil spot converges to the GT ellipse, thus forcing the network to produce elliptical inferences.

%%%%%%%%%%%%%%%%%%%%%%%%%%%%%%%%%%%%%%%%%%%%%%%%%%%%%%%%%%%%%%%%%%%%%%%%%%%%%%%%%%%%%%%%%%%%%%%%%%%%%%%%%%%%%%%%%%%%%%
% HalfCheetah and Ant
%%%%%%%%%%%%%%%%%%%%%%%%%%%%%%%%%%%%%%%%%%%%%%%%%%%%%%%%%%%%%%%%%%%%%%%%%%%%%%%%%%%%%%%%%%%%%%%%%%%%%%%%%%%%%%%%%%%%%%

\subsection{Coordinate Transformation Method(CoordTransNet)} 

For each set of points \( C, G, K, \) and \( O, \) the CoordTransNet algorithm calculates the Euclidean distance between each point and the central gaze point \( S. \) It identifies the point farthest from \( S \) and uses this distance as a reference to ensure the longest line segment becomes the diagonal of the coordinate axes. Subsequently, the CoordTransNet algorithm uses this reference to compute the transformed square \( C_{2}, G_{2}, K_{2}, \) and \( O_{2}, \) while also calculating the transformation ratios for each point. These ratios are then used to transform the coordinates of other gaze points. By employing the concept of linear interpolation, the algorithm ensures that the \( X \) and \( Y \) values of each gaze point undergo a uniform transformation resembling elastic deformation, rather than simple scaling. Consequently, the transformation ratios for each point vary. The CoordTransNet algorithm uses these ratios to compute the new positions of each point. All gaze points are divided into eight regions based on the \( X \)-axis, \( Y \)-axis, and the lines \( SC, SG, SK, \) and \( SO. \) The details of the transformation ratios are provided below, using the second quadrant as an example. The same procedures apply to the other quadrants, as elaborated in the code provided in the supplementary materials.
\begin{figure}[h]
	\centering
	\includegraphics [width=1.0\linewidth]{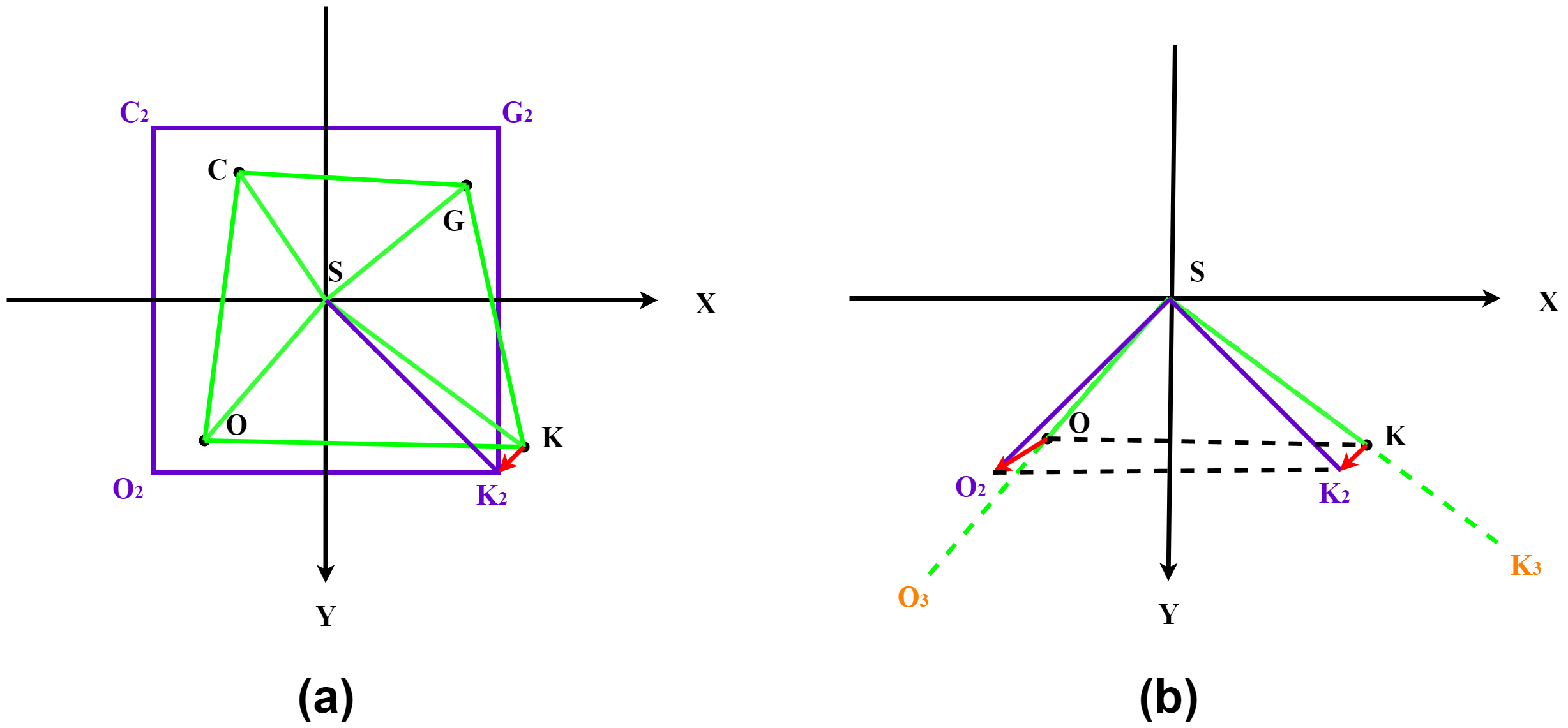}
	
	\caption{
      Coordinates Transformation
	}
	\label{fig:CoordinateTrans}
\end{figure}

% \[
% \text{if} \left( \left| \frac{{\text{y}}}{{\text{x}}} \right| > \left| \text{slopes}[K] \right| \right):
% \]
% \[
% \quad \text{change\_ratio\_x} = \delta_{K}'_{x}
% \]
% \[
% \quad \text{change\_ratio\_y} = \left( \frac{{\text{x} - [O][X]}}{{[K][X] - [O][X]}} \right) \cdot \delta_{K}'_{y} + \left( \frac{{[K][X] - \text{x}}}{{[K][X] - [O][X]}} \right) \cdot \delta_{O}'_{y}
% \]
% \[
% \text{elif} \left( \left| \frac{{\text{y}}}{{\text{x}}} \right| < \left| \text{slopes}[K] \right| \right):
% \]
% \[
% \quad \text{change\_ratio\_x} = \left( \frac{{\text{y} - [G][1]}}{{[K][Y] - [G][1]}} \right) \cdot \delta_{K}'_{x} + \left( \frac{{[K][Y] - \text{y}}}{{[K][Y] - [G][1]}} \right) \cdot \delta_{G}'_{x}
% \]
% \[
% \quad \text{change\_ratio\_y} = \delta_{K}'_{y}
% \]
% \[
% \text{else}:
% \]
% \[
% \quad \text{change\_ratio\_x} = \delta_{K}'_{x}
% \]
% \[
% \quad \text{change\_ratio\_y} = \delta_{K}'_{y}
% \]

% \text{if} \quad (x \neq 0) \quad \text{and} \quad \left|\frac{y}{x}\right| > \left| \text{slopes}[O]\right| :
if $x \neq 0$ and $\left|\frac{y}{x}\right| > \left|\text{slopes}[O]\right|$

\begin{align*}
\delta'_{O_{x}} &= \delta'_{O_{x}} \\
\delta'_{O_{y}} &= \left(\frac{x - O_{x}}{K_{x} - O_{x}}\right) \cdot \delta'_{K_{y}} + \left(\frac{K_{x} - x}{K_{x} - O_{x}}\right) \cdot \delta'_{O_{y}}
\end{align*}

% \text{elif} \quad (x \neq 0) \quad \text{and} \quad \left|\frac{y}{x}\right| < \left| \text{slopes}[O]\right| :
elif $x \neq 0$ and $\left|\frac{y}{x}\right| < \left|\text{slopes}[O]\right|$

\begin{align*}
\delta'_{O_{x}} &= \left(\frac{y - C_{y}}{O_{y} - C_{y}}\right) \cdot \delta'_{O_{x}} + \left(\frac{O_{y} - y}{O_{y} - C_{y}}\right) \cdot \delta'_{C_{x}} \\
\delta'_{O_{y}} &= \delta'_{O_{y}}
\end{align*}

\text{else} :

\begin{align*}
\delta'_{O_{x}} &= \delta'_{O_{x}} \\
\delta'_{O_{y}} &= \delta'_{O_{y}}
\end{align*}

% \delta'_{O_{x}} \quad \text{and} \quad \delta'_{O_{y}}: \\
$\delta'_{O_{x}} \text{ and } \delta'_{O_{y}}$

\text{Transformation ratios for  } O \text{ in the } x \text{ and } y \text{ directions.}

% \delta'_{K_{y}} \quad \text{and} \quad \delta'_{C_{x}}: \\
\[
\delta'_{K_{y}} \text{ and } \delta'_{C_{x}}
\]

\text{Transformation ratios for  } K \text{ and } C \text{ in the y and x directions.}

\begin{figure}[h]
	\centering
	\includegraphics [width=1.0\linewidth]{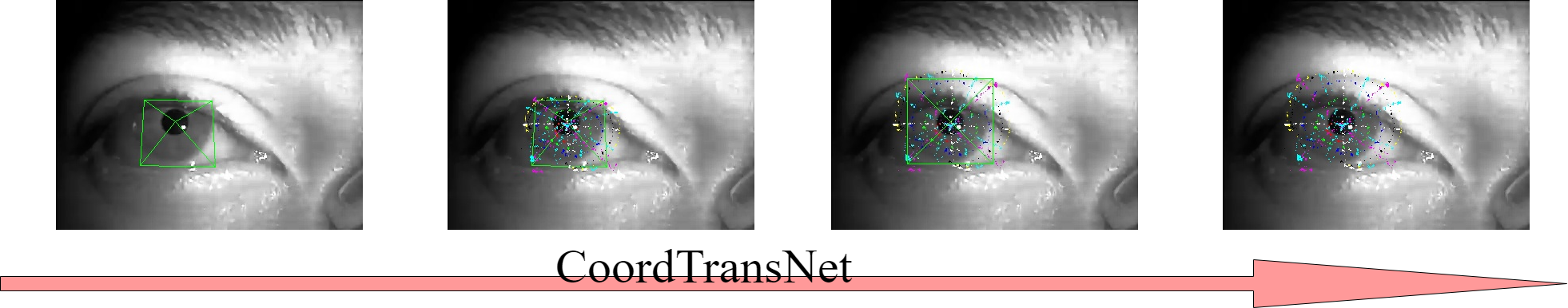}
	
	\caption{
      CoordTransNet
	}
	\label{fig:CoordTransNet}

\end{figure}

\subsection{Gaze Vector Generation Model(GVnet)}

We perform ellipse fitting using the coordinates of each transformed circular gaze point\cite{halir1998numerically} and linearly regress it to the standard gaze view. Here, we employ the method of least squares for fitting. Due to precise semantic segmentation performed earlier, a significant reduction in outliers has been achieved, akin to a purification process prior to fitting. In case of any remaining outlier interference, filtering conditions can be set to remove them.

The method involves fitting the ellipse of the pupil using the least squares approach, by minimizing the sum of squared distances from the semantic segmentation edge data points to the fitted ellipse\cite{fuhl2020learning}, in order to determine the parameters of the ellipse. The method is as follows:

Firstly, based on the general equation of an ellipse:
\begin{equation}
    F(W;X) = Ax^2 + Bxy + Cy^2 + Dx + Ey + F = 0
    \label{eq:ellipse_equation}
\end{equation}

Let:
\begin{equation}
    W = \begin{bmatrix} A , B , C , D , E , F \end{bmatrix}^T
\end{equation}
\begin{equation}
    X = \begin{bmatrix} x^2 , xy , y^2 , x , y , 1 \end{bmatrix}^T
\end{equation}

The objective function to be optimized can be determined as:
\begin{equation}
    \min E(W) = ||DW||^2
    \label{eq:objective_function}
\end{equation}

where $D$ is an $n \times 6$ matrix $[X_1 \ X_2 \ \ldots \ X_n]^T$ and $||W^TX||^2 = W^TXX^TW$.

subject to $W^THW > 0$ where
\[
H = \begin{bmatrix}
0 & 0 & 2 & 0 & 0 & 0 \\
0 & -1 & 0 & 0 & 0 & 0 \\
2 & 0 & 0 & 0 & 0 & 0 \\
0 & 0 & 0 & 0 & 0 & 0 \\
0 & 0 & 0 & 0 & 0 & 0 \\
0 & 0 & 0 & 0 & 0 & 0 \\
\end{bmatrix}
\]

In equation \ref{eq:ellipse_equation}, $W^THW > 0$ represents the constraint of ellipse parameters $4AC - B^2 > 0$. Since $W$ is a coefficient vector, enlarging or reducing the coefficients simultaneously in the ellipse equation does not affect the result. Thus, when $||W^TX||^2 = 0$, any $W' = \alpha W (\alpha > 0)$ satisfies the condition. Therefore, the constraint equation can be modified to:
\begin{equation}
    W^THW = 1
    \label{eq:constraint_equation}
\end{equation}

Based on equations \ref{eq:objective_function} and \ref{eq:constraint_equation}, the Lagrangian function can be constructed as:
\begin{equation}
    L(W,\lambda) = ||DW||^2 - \lambda(W^THW - 1)
    \label{eq:lagrangian_function}
\end{equation}

Taking the derivative of equation \ref{eq:lagrangian_function} with respect to $W$ and setting it equal to zero yields:
\begin{equation}
    D^T DW - \lambda HW = 0
    \label{eq:derivative_zero}
\end{equation}

At this point, let $S = D^T D$, then $SW = \lambda HW$. Since $S$ is a positive definite matrix, multiplying $S^{-1}$ to both sides of the equation results in:
\begin{equation}
    S^{-1} HW = \frac{1}{\lambda} W
    \label{eq:solution_equation}
\end{equation}

In equation \ref{eq:solution_equation}, $S^{-1}$ and $H$ are known, and solving this equation is equivalent to solving for the eigenvectors, ultimately yielding the equation of the ellipse.

The GVnet model includes the following key components: a self-attention layer is used to capture the relationship in the input data. Second, the input data is processed through multiple hidden layers, namely fully connected layers. Since the input data in some datasets are not all positive, the LeakyRelu activation function is used in the hidden layer. In addition, a Dropout layer is added to prevent the model from overfitting.

\begin{figure}
	\centering
	\includegraphics [width=1.0\linewidth]{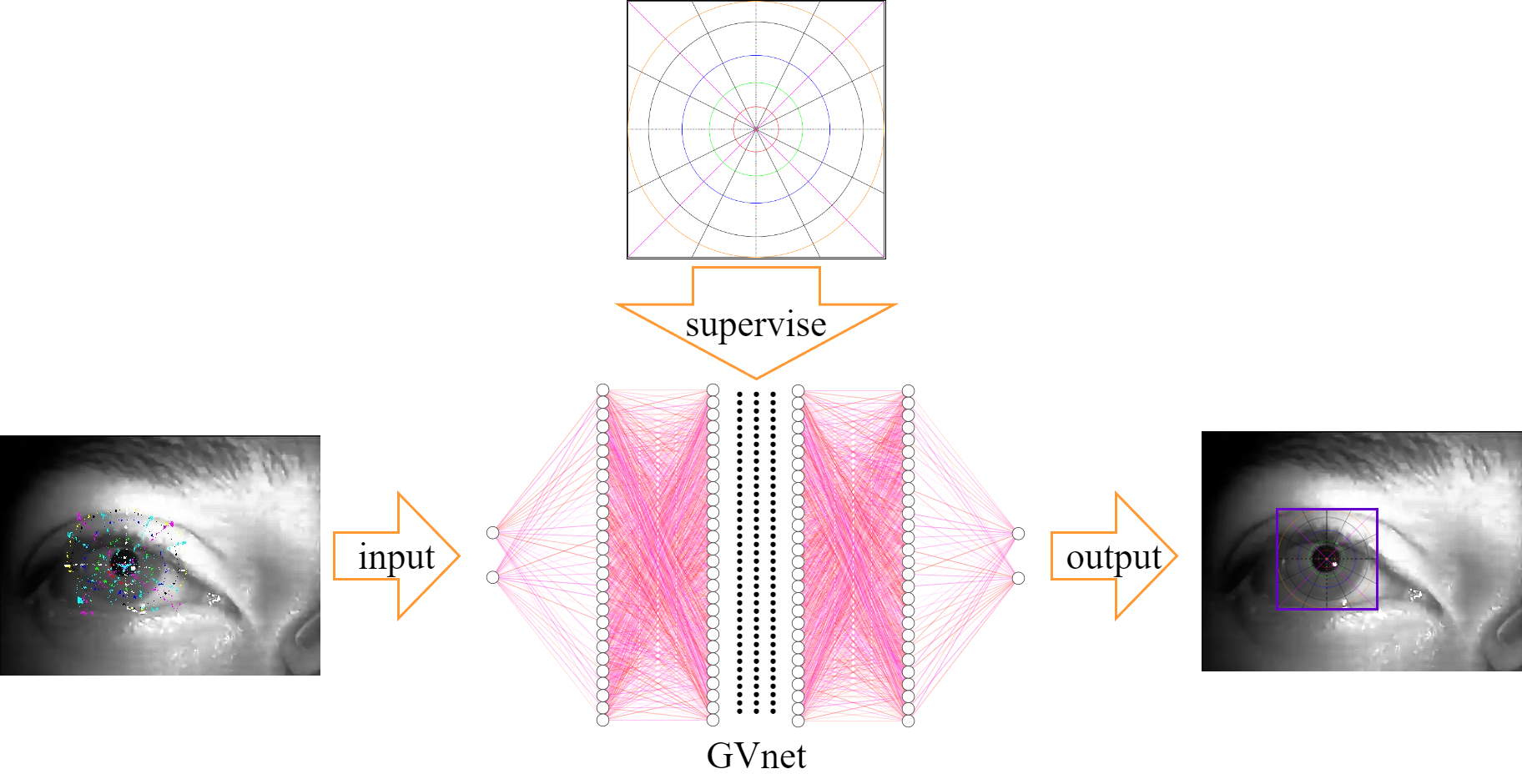}
	
	\caption{
      GVnet model
	}
	\label{fig:GVnet}
\end{figure}

%% file: sections/dataset.tex
\begin{figure}[h]
	\centering
	\includegraphics [width=1.0\linewidth]{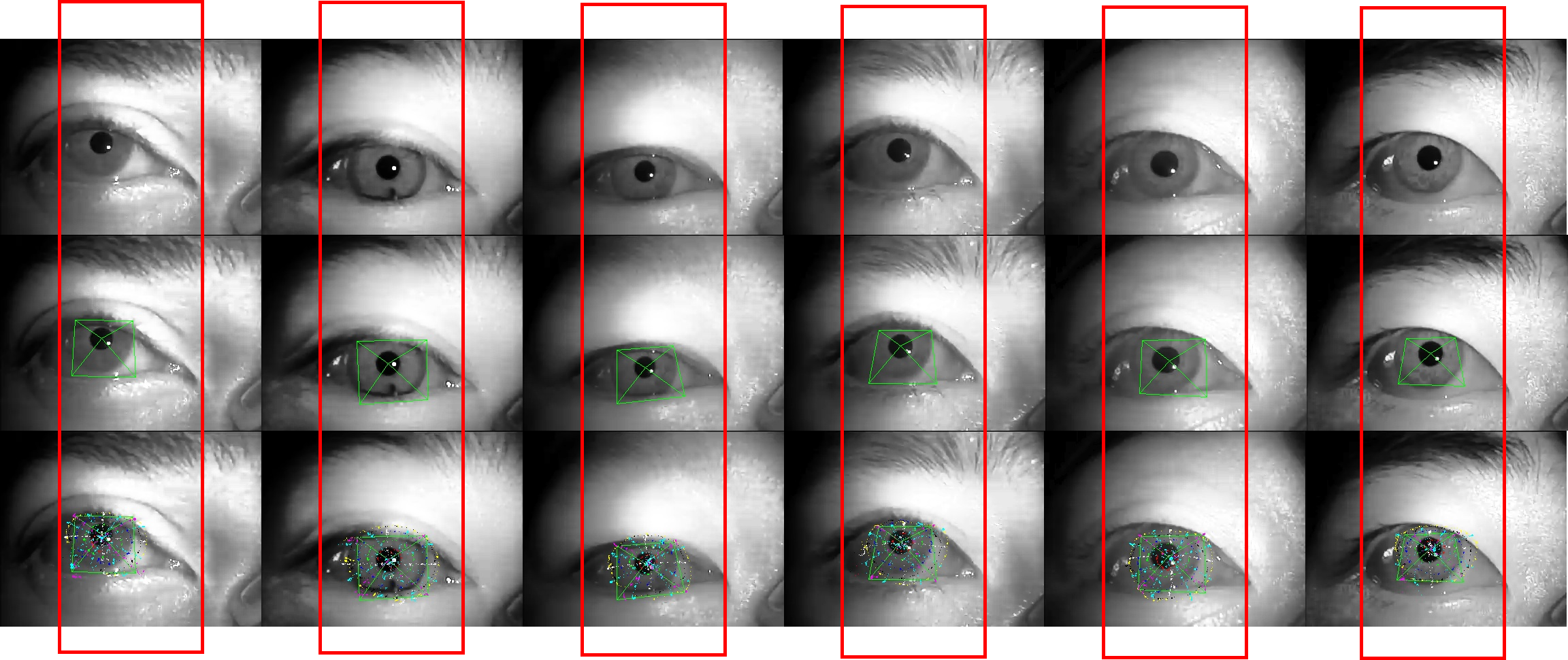}
	
	\caption{
      GazeTrack Dataset
	}
	\label{fig:dataset}
\end{figure}
\section{Dataset}
To address the lack of precise training data for eye tracking, we invented standard gaze views and collected a new gaze dataset (GazeTrack) using laser positioning and eye trackers. GazeTrack comprises dynamic eye gaze data from 47 volunteers of various ages, genders, and ethnicities. These data were captured using cameras on the DG3 eye tracker and annotated with precise pupil coordinates using semantic segmentation and deformation transformation. The DG3 camera captures data from the left eye, with a resolution of 384*288 pixels and a frame rate of 60FPS. Unlike previous datasets, we captured multi-angle images, including points, lines, and circles, with distinct features and broad applicability.For more details about the dataset, such as the collection method and environment, category definition, data distribution, etc., please refer to the supplementary materials.

%% file: sections/experiments.tex
\section{Experiments}
The three experimental results of semantic segmentation, coordinate transformation, and vector generation will be described below.For more details about the experiments please refer to the supplementary materials.

\subsection{Pupil semantic segmentation}

\textbf{Experimental Setup} We trained the U-ResAtt semantic segmentation model on the LPW dataset with precise annotations and filtering, using a training-validation ratio of 4:1. We initialized a UNet with 32 feature channels. Adam optimizer was employed to optimize the model parameters, with a learning rate set to 1e-4. A learning rate scheduler, `scheduler`, was utilized to dynamically adjust the learning rate when the validation loss no longer decreased. We added a regularization term based on ellipse fitting error to the binary cross-entropy loss. Training was stopped early if the validation loss did not decrease for five consecutive rounds.

\textbf{Results} 
Table \ref{mean-accuracy} presents the performance results of pupil detection for 24 participants in the ExCuSe dataset using different algorithms under five-pixel error thresholds (indicating the maximum allowable deviation between predicted and actual pupil positions, measured in pixels). The results are expressed as percentages.

\begin{table}[h]
  \centering
  \begin{tabular}{lc}
    \toprule
    Model                   & Mean Accuracy \\
    \midrule
    Swirski\shortcite{swirski2012robust}                   & 38.23      \\
    ExCuSe\shortcite{fuhl2015excuse}                    & 58.31      \\
    EISe\shortcite{fuhl2016else}                      & 68.20      \\
    UNet (base)\shortcite{ronneberger2015u}               & 89.59      \\
    UNet (Convex Shape)       & 89.78      \\
    U-ResAtt (Ellipse Fit Error) & 90.27  \\
    \bottomrule
  \end{tabular}
    \caption{Mean Accuracy for each model}
    \label{mean-accuracy}
\end{table}

\subsection{Coordinates Transform Accuracy} 

\textbf{Experimental Setup} We use the collected accurate dataset GazeTrack for coordinate transformation, which can achieve millisecond-level response on NVIDIA TITAN V.Since this method involves pure mathematical logic, there is no parameter setting. I will perform ablation analysis to analyze the superiority of this method.

\textbf{Results} 
Table \ref{mean-acc} show that coordinate transformation reduces pixel-level loss.

\begin{table}[h]
  \centering
  \begin{tabular}{lc}
    \toprule
    Method                   & Pixel Error \\
    \midrule
    without CoordTransNet            & 7.20      \\
    CoordTransNet                    & 3.84      \\
    \bottomrule
  \end{tabular}
    \caption{Ablation Experiment}
    \label{mean-acc}
\end{table}

\subsection{Gaze Vector Generation Accuracy}

\textbf{Experimental Setup} An early stopping strategy is added to prevent the model from overfitting. The model loss function is the mean square error (MSE) loss function, and the Adam optimizer is used to update the model parameters.

\textbf{Results} 
In order to prove the vector generation efficiency of this model, that is, the computational complexity, we use 1000 data input and output calculation time cost on a single core of Intel Core i5 on the desktop.The method for calculating the error in the center of the eyeball is the Mean Absolute Error In Pixel (MAE-Pixel), which represents the average absolute distance between the estimated center point and the actual center point, measured in pixels. The method for calculating the error in gaze vectors is the Mean Absolute Error In Degree (MAE-Degree), which represents the average absolute angle between the estimated gaze vector and the actual gaze vector, measured in degrees. All errors are reported as their respective averages.Regarding the computational complexity of the models, the time cost is computed based on 1000 input-output data pairs, measured in seconds. Table presents the results of different models and algorithms, where W1 indicates a sliding window size of 1, and so forth.

\begin{table*}[t]
\centering
\begin{tabular}{lcccccc}
\toprule
Model   & \multicolumn{3}{c}{Gaze Vector Error (degrees)} & \multicolumn{3}{c}{Execution Time (seconds)} \\
\cmidrule(lr){2-4} \cmidrule(lr){5-7}
        & W1  & W10 & W20  & W1  & W10 & W20  \\
\midrule
NN30LM\shortcite{fuhl2020neural}  & 12.57 & 2.58 & 1.82 & 5.00 & 0.50 & 0.47 \\
NN50BR\shortcite{fuhl2020neural}  & 12.33 & 1.73 & 1.02 & 5.00 & 0.34 & 0.32 \\
T5Bag\shortcite{fuhl2020neural}   & 11.37 & 2.81 & 3.16 & 5.00 & 1.48 & 2.15 \\
SVMLIN\shortcite{fuhl2020neural}  & 13.67 & 5.02 & 5.08 & 5.00 & 1.03 & 1.03 \\
SVMGAU\shortcite{fuhl2020neural}  & 12.40 & 11.45 & 19.64 & 5.00 & 1.08 & 1.49 \\
GVnet   & 9.37  & 1.35 & 0.94 & 3.93 & 1.01 & 0.75 \\
\bottomrule
\end{tabular}
\caption{Performance Comparison of Different Models}
\label{tab:model-performance}
\end{table*}

Next, I will compare the average error of the gaze vector generated by the entire GazeTrack method with other algorithms on third-party data.
\begin{table}[h]
  \centering
  \begin{tabular}{lc}
    \toprule
    Method              & Gaze Vector Error (degrees) \\
    \midrule
    Swirski\shortcite{swirski2013fully}                      & 7.82      \\
    Fuhl\shortcite{fuhl2020neural}                         & 5.69      \\
    Dierkes\shortcite{dierkes2018novel}                      & 5.39      \\
    3dGaze\shortcite{lu2022neural}                       & 4.38      \\
    GazeTrack                    & 2.84      \\
    \bottomrule
  \end{tabular}
    \caption{Results on Neural 3D Gaze}
    \label{mean-3D}
\end{table}

%% file: sections/conclusion.tex
\section{Conclusion}

We addresses the current limitations in gaze accuracy for spatial computing in virtual and augmented reality applications. A gaze collection framework called GazeTrack was developed, along with high-precision equipment, to gather a benchmark dataset encompassing diverse ethnicities, ages, and visual acuity conditions. A novel shape error regularization method was proposed to improve pupil ellipse fitting accuracy, and training was conducted on open-source datasets to enhance semantic segmentation and pupil position prediction accuracy. Additionally, a novel coordinate transformation method, similar to paper unfolding, was introduced to accurately predict gaze vectors on the GazeTrack dataset. Finally, a gaze vector generation model was developed, achieving reduced gaze angle error with lower computational complexity compared to existing methods.

%% file: sections/checklist.tex
\title{Checklist}

\section{Reproducibility Checklist}

This paper:
\begin{itemize}
    \item Includes a conceptual outline and/or pseudocode description of AI methods introduced (\red{yes}/partial/no/NA)
    \item Clearly delineates statements that are opinions, hypothesis, and speculation from objective facts and results (\red{yes}/no)
    \item Provides well marked pedagogical references for less-familiare readers to gain background necessary to replicate the paper (\red{yes}/no)
\end{itemize}

\noindent{Does this paper make theoretical contributions? (\red{yes}/no) }

If yes, please complete the list below.
\begin{itemize}
    \item All assumptions and restrictions are stated clearly and formally. (\red{yes}/partial/no)
    \item All novel claims are stated formally (e.g., in theorem statements). (\red{yes}/partial/no)
    \item Proofs of all novel claims are included. (\red{yes}/partial/no)
    \item Proof sketches or intuitions are given for complex and/or novel results. (\red{yes}/partial/no)
    \item Appropriate citations to theoretical tools used are given. (\red{yes}/partial/no)
    \item All theoretical claims are demonstrated empirically to hold. (\red{yes}/partial/no/NA)
    \item All experimental code used to eliminate or disprove claims is included. (\red{yes}/no/NA)
\end{itemize}

\noindent{Does this paper rely on one or more datasets? (\red{yes}/no) }

If yes, please complete the list below.
\begin{itemize}
    \item A motivation is given for why the experiments are conducted on the selected datasets (\red{yes}/partial/no/NA)
    \item All novel datasets introduced in this paper are included in a data appendix. (\red{yes}/partial/no/NA)
    \item All novel datasets introduced in this paper will be made publicly available upon publication of the paper with a license that allows free usage for research purposes. (\red{yes}/partial/no/NA)
    \item All datasets drawn from the existing literature (potentially including authors’ own previously published work) are accompanied by appropriate citations. (\red{yes}/no/NA)
    \item All datasets drawn from the existing literature (potentially including authors’ own previously published work) are publicly available. (\red{yes}/partial/no/NA)
    \item All datasets that are not publicly available are described in detail, with explanation why publicly available alternatives are not scientifically satisficing. (\red{yes}/partial/no/NA)
\end{itemize}

\noindent{Does this paper include computational experiments? (\red{yes}/no) }

If yes, please complete the list below.
\begin{itemize}
    \item Any code required for pre-processing data is included in the appendix. (\red{yes}/partial/no).
    \item All source code required for conducting and analyzing the experiments is included in a code appendix. (\red{yes}/partial/no)
    \item All source code required for conducting and analyzing the experiments will be made publicly available upon publication of the paper with a license that allows free usage for research purposes. (\red{yes}/partial/no)
    \item All source code implementing new methods have comments detailing the implementation, with references to the paper where each step comes from (\red{yes}/partial/no)
    \item If an algorithm depends on randomness, then the method used for setting seeds is described in a way sufficient to allow replication of results. (\red{yes}/partial/no/NA)
    \item This paper specifies the computing infrastructure used for running experiments (hardware and software), including GPU/CPU models; amount of memory; operating system; names and versions of relevant software libraries and frameworks. (\red{yes}/partial/no)
    \item This paper formally describes evaluation metrics used and explains the motivation for choosing these metrics. (\red{yes}/partial/no)
    \item This paper states the number of algorithm runs used to compute each reported result. (\red{yes}/no)
    \item Analysis of experiments goes beyond single-dimensional summaries of performance (e.g., average; median) to include measures of variation, confidence, or other distributional information. (\red{yes}/no)
    \item The significance of any improvement or decrease in performance is judged using appropriate statistical tests (e.g., Wilcoxon signed-rank). (\red{yes}/partial/no)
    \item This paper lists all final (hyper-)parameters used for each model/algorithm in the paper’s experiments. (\red{yes}/partial/no/NA)
    \item This paper states the number and range of values tried per (hyper-) parameter during development of the paper, along with the criterion used for selecting the final parameter setting. (\red{yes}/partial/no/NA)
\end{itemize}

% \end{document}